\documentclass[11pt,a4paper]{article}
\usepackage[hyperref]{tacl2018} %
\usepackage{times,latexsym}
\usepackage{url}
\usepackage[T1]{fontenc}

\taclfinalcopy

\usepackage{xspace,mfirstuc,tabulary}
\usepackage{amsmath,bm}
\usepackage{multirow}
\usepackage{color}
\usepackage{xcolor}
\usepackage{multicol}
\usepackage{tabularx}
\usepackage{booktabs}
\usepackage{pbox}
\usepackage{framed}
\usepackage{array}
\usepackage{enumitem}
\usepackage{chngpage}
\usepackage{threeparttable}
\usepackage{dsfont}
\usepackage{graphicx}
\usepackage{enumitem}

\newcommand{\eg}{{\sl e.g.}}
\newcommand{\ie}{{\sl i.e.}}
\newcommand{\vs}{{\sl vs. }}
\newcommand{\hhide}[1]{}

\iftaclfinal

\else

\fi

\newif\iftaclinstructions
\taclinstructionsfalse
\iftaclinstructions
\renewcommand{\confidential}{}
\renewcommand{\anonsubtext}{(No author info supplied here, for consistency with
TACL-submission anonymization requirements)}
\fi

\title{DREAM: A Challenge Dataset and Models for \\Dialogue-Based Reading Comprehension}
\author{
 Kai Sun\textsuperscript{$\spadesuit$}\Thanks{This work was done when the author was an intern at Tencent AI Lab, Bellevue, WA.} ~~ Dian Yu\textsuperscript{$\heartsuit$} ~~ Jianshu Chen\textsuperscript{$\heartsuit$} ~~ Dong Yu\textsuperscript{$\heartsuit$} ~~ Yejin Choi\textsuperscript{$\diamondsuit$ $\clubsuit$} ~~Claire Cardie\textsuperscript{$\spadesuit$} \\
 \textsuperscript{$\spadesuit$}Cornell University, Ithaca, NY, USA  \\
 \textsuperscript{$\heartsuit$}Tencent AI Lab, Bellevue, WA, USA\\
 \textsuperscript{$\diamondsuit$}University of Washington, Seattle, WA, USA \\
  \textsuperscript{$\clubsuit$}Allen Institute for Artificial Intelligence, Seattle, WA, USA \\
  {\sf  } \\
}

\date{}

\begin{document}
\maketitle

\begin{abstract}

We present DREAM, the first dialogue-based multiple-choice reading comprehension dataset. Collected from English-as-a-foreign-language examinations designed by human experts to evaluate the comprehension level of Chinese learners of English, our dataset contains 10,197 multiple-choice questions for 6,444 dialogues. In contrast to existing reading comprehension datasets, DREAM is the first to focus on in-depth multi-turn multi-party dialogue understanding. DREAM is likely to present significant challenges for existing reading comprehension systems: 
$84\%$ of answers are non-extractive, $85\%$ of questions require reasoning beyond a single sentence, and $34\%$ of questions also involve commonsense knowledge.

We apply several popular neural reading comprehension models that primarily exploit surface information within the text and find them to, at best, just barely outperform a rule-based approach. We next investigate the effects of incorporating dialogue structure and different kinds of general world knowledge into both rule-based and (neural and non-neural) machine learning-based reading comprehension models. Experimental results on the DREAM dataset show the effectiveness of dialogue structure and general world knowledge. DREAM will be available at \url{https://dataset.org/dream/}.

\end{abstract}
\section{Introduction}

\begin{table}[h!]
\centering
\footnotesize
\begin{tabular}{p{0.2cm}p{6.5cm}}
\toprule
\textbf{Dialogue$~$1$~$(D1)}                               &             \\
\midrule
\textbf{W}: & Tom, look at your shoes. How dirty they are! You must clean them. \\
\textbf{M}: & Oh, mum, I just cleaned them yesterday. \\
\textbf{W}: & They are dirty now. You must clean them again. \\
\textbf{M}: & I do not want to clean them today. Even if I clean them today, they will get dirty again tomorrow.\\
\textbf{W}: & All right, then. \\
\textbf{M}: & Mum, give me something to eat, please. \\
\textbf{W}: & You had your breakfast in the morning, Tom, and you had lunch at school.\\
\textbf{M}: & I am hungry again.\\
\textbf{W}: & Oh, hungry? But if I give you something to eat today, you will be hungry again tomorrow.\\
\midrule
\textbf{Q1}                            & Why did the woman say that she wouldn't give him anything to eat?             \\
A.   &  Because his mother wants to correct his bad habit.$\star$  \\
B.   &  Because he had lunch at school.                            \\
C.   &  Because his mother wants to leave him hungry.              \\
\bottomrule
\end{tabular}
\caption{A sample DREAM problem that requires general world knowledge ($\star$: the correct answer option).}
\label{tab:sample1}
\end{table}

Recently a significant amount of research has focused on the construction of large-scale multiple-choice~\cite{lai2017race,khashabi2018looking,ostermann2018semeval} and extractive ~\cite{hermann2015teaching,hill2015goldilocks,rajpurkar2016squad,trischler2017newsqa} reading comprehension datasets (Section~\ref{sec:related}). Source documents
in these datasets have generally been drawn from formal written texts such as news, fiction, and Wikipedia articles, which are commonly considered well-written, accurate, and neutral.

With the goal of advancing research in machine reading comprehension and facilitating dialogue understanding, we construct and present in this paper DREAM --- the first multiple-choice \textbf{D}ialogue-based \textbf{REA}ding comprehension exa\textbf{M}ination dataset. We collect 10,197 questions for 6,444 multi-turn multi-party dialogues from English language exams that are carefully designed by educational experts (\eg, English teachers) to assess the comprehension level of Chinese learners of English. 
Each question is associated with three answer options, exactly one of which is correct. 
(See Table~\ref{tab:sample1} for an example.) DREAM covers a variety of topics and scenarios in daily life such as conversations on the street, on the phone, in a classroom or library, at the airport or the office or a shop (Section~\ref{sec:data}).%

Based on our analysis of DREAM, we argue that dialogue-based reading comprehension is at least as difficult as existing non-conversational counterparts. In particular, answering $\textbf{34\%}$ of DREAM questions requires unspoken commonsense knowledge, \eg, unspoken scene information. This might be due to the nature of dialogues: for efficient oral communication, people rarely state obvious explicit world knowledge~\cite{forbes2017verb} such as \emph{``Christmas Day is celebrated on December 25th''}. Understanding the social implications of an utterance as well as inferring a speaker's intentions is also regularly required for answering dialogue-based questions. The dialogue content in Table~\ref{tab:sample1}, for example, is itself insufficient for readers to recognize the intention of the female speaker (W) in the first question (Q$1$). 
However, world knowledge is rarely considered in state-of-the-art reading comprehension models~\cite{tay2018multi,wang2018co}. 

Moreover, dialogue-based questions can cover information imparted across multiple turns involving multiple speakers. In DREAM, approximately 
$\textbf{85\%}$ of questions can only be answered by considering the information from multiple sentences. For example, to answer Q$1$ in Table~\ref{tab:sample} regarding the date of birth of the male speaker (M), the supporting sentences (in bold) include \emph{``You know, tomorrow is Christmas Day''} from the female speaker and \emph{``\dots I am more than excited about my birthday, which will come in two days''} from the male speaker. %
Compared to ``multiple-sentence questions'' in traditional reading comprehension datasets, DREAM further requires an understanding of the turn-based structure of dialogue, \eg \ for aligning utterances with their corresponding speakers.

As only $\textbf{16\%}$ of correct answers are text spans from the source documents, we primarily explore rule-based methods and state-of-the-art neural models designed for multiple-choice reading comprehension (Section~\ref{sec:methods}). 
We find first that neural models designed for non-dialogue-based reading comprehension~\cite{chen2016thorough,dhingra2017gated,wang2018co} do not fare well: the highest achieved accuracy is $45.5\%$, only slightly better than the accuracy of a simple lexical baseline~\cite{richardson2013mctest} ($44.6\%$).
For the most part, these models fundamentally exploit only surface-level information from the source documents. Considering the above-mentioned challenges, however, we hypothesize that incorporating general world knowledge and aspects of the dialogue structure would allow a better understanding of the dialogues. As a result, we modify our baseline systems to include (1) general world knowledge in the form of such as ConceptNet relations~\cite{speer2017conceptnet} and a pre-trained language model~\cite{radfordimproving}, and (2) speaker information for each utterance. Experiments show the effectiveness of these factors on the lexical baselines as well as neural and non-neural machine learning approaches: we acquire up to $11.9\%$ absolute gain in accuracy compared to the highest performance achieved by the state-of-the-art reading comprehension model~\cite{wang2018co} that mainly relies on explicit surface-level information in the text (Section~\ref{sec:experiment}).

Finally, we see a significant gap between the best automated approach ($59.5\%$) and human ceiling performance ($98.6\%$) on the DREAM dataset. This provides yet additional evidence that dialogue-based reading comprehension is a very challenging task. 
We hope that it also inspires the research community to develop methods for the dialogue-based reading comprehension task.
\section{Related Work}
\label{sec:related}
We divide reading comprehension datasets into three categories based on the types of answers.

\begin{table*}[ht!]
\centering
\footnotesize
\begin{tabular}{lccccc}
\toprule
                    & \textbf{SQuAD}   &\textbf{NarrativeQA}   &\textbf{CoQA}  &\textbf{RACE}   &\textbf{DREAM (this work)}  \\
\midrule            
Answer type               & extractive      & abstractive    & abstractive   & multiple-choice  & multiple-choice \\
Source document type   &  written text   & written text   & written text  &  written text    & dialogue    \\
\# of source documents &  536            & 1,572          &  8,399        &   27,933         &  6,444 \\
Average answer length  & 3.2             & 4.7     & 2.7          & 5.3            & 5.3             \\

\midrule
Extractive (\%)     & 100.0           &73.6    & 66.8          & 13.0           & 16.3           \\
Abstractive (\%)    & 0.0             &26.4    & 33.2          & 87.0           & 83.7           \\
\bottomrule
\end{tabular}
\caption{Distribution of answer (or correct answer option) types in three kinds of reading comprehension datasets. Statistics of other datasets come from~\newcite{reddy2018coqa},~\newcite{kovcisky2018narrativeqa}, and~\newcite{lai2017race}.}
\label{tab:related:answer}
\end{table*}

\subsection{Extractive and Abstractive Datasets}
In recent years, we have seen increased interest in large-scale cloze/span-based reading comprehension dataset construction~\cite{hermann2015teaching,hill2015goldilocks,onishi2016did,rajpurkar2016squad,bajgar2016embracing,nguyen2016ms,trischler2017newsqa,triviaQA,choi2018quac}. We regard them as extractive since candidate answers are usually short spans from source documents. State-of-the-art neural models with attention mechanisms already achieve very high performance based on local lexical information. Recently researchers work on the construction of spoken span-based datasets~\cite{ODSQA,li2018spoken} by applying text-to-speech technologies or recruiting human speakers based on formal written document-based datasets such as \textbf{SQuAD}~\cite{rajpurkar2016squad}. Some span-based conversation datasets are constructed from a relatively small size of dialogues from TV shows~\cite{chen2016character,ma2018challenging}. 

Considering the limitations in extractive datasets, answers in abstractive datasets such as MS MARCO~\cite{nguyen2016ms}, SearchQA~\cite{dunn2017searchqa}, and \textbf{NarrativeQA}~\cite{kovcisky2018narrativeqa} are human generated based on source documents or summaries. Concurrently, there is a growing interest in conversational question answering such as \textbf{CoQA}~\cite{reddy2018coqa}. Since annotators tend to copy spans as answers~\cite{reddy2018coqa}, the majority of answers are still extractive in these datasets (Table~\ref{tab:related:answer}). Compared to the datasets mentioned above, most of the correct answer options ($\textbf{83.7\%}$) in DREAM are free-form text.

\subsection{Multiple-Choice Datasets}
We primarily discuss the multiple-choice datasets in which answer options are not restricted to extractive text spans in the given document. Instead, most of the correct answer options are abstractive (Table~\ref{tab:related:answer}). Multiple-choice datasets involve extensive human involvement for problem generation during crowdsourcing (\ie, questions, correct answer option, and distractors). Besides surface matching, a significant portion of questions require multiple-sentence reasoning and external knowledge~\cite{richardson2013mctest,mostafazadeh2016corpus,khashabi2018looking,ostermann2018semeval}. 

Besides crowdsourcing, some datasets are collected from examinations designed by educational experts~\cite{penas2014overview,shibuki2014overview,tseng2016towards, clark2016combining,lai2017race}, which aim to test human examinees. There are various types of complicated questions such as math word problems, summarization, logical reasoning, and sentiment analysis. Since we can adopt more objective evaluation criteria such as accuracy, these questions are usually easy to grade. Besides, questions from examinations are generally clean and high-quality. Therefore, human performance ceiling on this kind of datasets is much higher (\eg, $94.5\%$ on \textbf{RACE}~\cite{lai2017race} and $98.6\%$ on DREAM in accuracy) than that of datasets built by crowdsourcing.

In comparison, we present the first multiple-choice \emph{\bf dialogue-based} dataset from examinations that contains a large percentage of questions that require multiple sentence inference. To the best of our knowledge, DREAM also contains the largest number of questions involving \emph{\bf commonsense reasoning} compared to other examination datasets. 
\section{Data}
\label{sec:data}

\begin{table}[!htb]
\centering
\footnotesize
\begin{tabular}{p{0.2cm}p{6.5cm}}
\toprule
\textbf{Dialogue$~$2$~$(D2)}                               &            \\
\midrule
\textbf{W}:   &  Hey, Mike. Where have you been? I didn't see you around these days?          \\
\textbf{M}:   &  I was hiding in my office. My boss gave me loads of work to do, and I tried to finish it before my birthday. Anyway, I am done now. Thank goodness! How is everything going with you?         \\
\textbf{W}:   &  I'm quite well. \textbf{You know, tomorrow is Christmas Day.} Do you have any plans?    \\
\textbf{M}:   &  \textbf{Well, to tell you the truth, I am more than excited about my birthday, which will come in two days.} I am going to visit my parents-in-law with my wife.         \\
\textbf{W}:   &   Wow, sounds great. \\
\textbf{M}:   &   Definitely! This is my first time to spend my birthday with them.  \\
\textbf{W}:   &   Do they live far away from here?  \\
\textbf{M}:   &   A little bit. We planned to take the train, but considering the travel peak, my wife strongly suggested that we go to the airport right after we finish our work this afternoon. How about you? What's your holiday plan?\\
\textbf{W}:   &   Well, our situations are just the opposite. My parents-in-law will come to my house, and they wish to stay at home and have a quiet Christmas Day. So I have to call my friends to cancel our party that will be held at my house. \\
\textbf{M}:   &   You'll experience a quite different and lovely holiday. Enjoy your Christmas! \\
\textbf{W}:   &   Thanks, the same to you! \\
\midrule
\textbf{Q1}                            & What is the date of the man's birthday?           \\
A.   &  25th, December.            \\
B.   &  26th, December.$\star$          \\
C.   &  27th, December.           \\
\textbf{Q2}                             & How will the man go to his wife's parents' home?            \\
A.   & By train.           \\
B.   & By bus.            \\
C.   & By plane.$\star$          \\
\textbf{Q3}                              & What is the probable relationship between the two speakers?            \\
A.   &  Husband and wife.            \\
B.   &  Friends.$\star$           \\
C.   &  Parent-in-law and son-in-law.            \\
\bottomrule
\end{tabular}
\caption{A complete sample DREAM problem ($\star$: the correct answer option). }
\label{tab:sample}
\end{table}

In this section, we describe how we construct DREAM (Section~\ref{collection}) and provide a detailed analysis of this dataset (Section~\ref{analysis}).

\subsection{Collection Methodology}
\label{collection}

We collect dialogue-based comprehension problems from a variety of English language exams (including practice exams) such as National College Entrance Examination, College English Test, and Public English Test\footnote{We list all the websites used for data collection in the released dataset.}, which are designed by human experts to assess either the listening or reading comprehension level of Chinese English learners in high schools and colleges (aged $12$-$22$). All the problems in DREAM are freely accessible online for public usage. Each problem consists of a dialogue and a series of multiple-choice questions. To ensure every question is associated with exactly three answer options, we drop wrong option(s) randomly for questions with more than three options. We remove duplicate problems and randomly split the data at the problem level, with $60\%$ train, $20\%$ development, and $20\%$ test.

\subsection{Data Analysis}
\label{analysis}

We summarize the statistics of DREAM in Table~\ref{tab:data:overall} and data split in Table~\ref{tab:data:stat}. Compared to existing datasets built from formal written texts, the vocabulary size is relatively small since spoken English by its nature makes greater use of high-frequency words and needs a smaller vocabulary for efficient real-time communication ~\cite{nation2006large}.

\begin{table}[htbp!]
\centering
\footnotesize
\begin{tabular}{lc}
\toprule
\textbf{Metric} & \textbf{Value}   \\
\midrule

\# of answer options per question          & 3  \\
\# of turns                         &  30,183 \\
Avg./Max. \# of questions per dialogue & 1.6 / 10 \\
Avg./Max. \# of speakers per dialogue  & 2.0 / 7  \\
Avg./Max. \# of turns per dialogue  &  4.7 / 48  \\ 
Avg./Max. option length (in tokens)   &  5.3 / 21  \\
Avg./Max. question length (in tokens)  &  8.6 / 24  \\
Avg./Max. dialogue length (in tokens)  &  85.9 / 1,290  \\
vocabulary size                     &  13,037  \\
\bottomrule
\end{tabular}
\caption{The overall statistics of DREAM. A turn is defined as an uninterrupted stream of speech from one speaker in a dialogue.}
\label{tab:data:overall}
\end{table}

\begin{table}[htbp!]
\centering
\footnotesize
\begin{tabular}{lcccc}
\toprule
 & \bf Train & \bf Dev & \bf Test & \bf All  \\
\midrule
\# of dialogues & 3,869 & 1,288 & 1,287 & 6,444 \\
\# of questions & 6,116 & 2,040  & 2,041 & 10,197 \\
\bottomrule
\end{tabular}
\caption{The separation of the training, development, and test sets in DREAM.}
\label{tab:data:stat}
\end{table}

We categorize questions into two main categories according to the types of knowledge required to answer them: \textit{matching} and \textit{reasoning}.

\begin{itemize}%
\item \textbf{Matching} A question is entailed or paraphrased by exactly one sentence in a dialogue. The answer can be extracted from the same sentence. For example, we can easily verify the correctness of the question-answer pair (\emph{``What kind of room does the man want to rent?''}, \emph{``A two-bedroom apartment.''}) based on the sentence \emph{``\textbf{M}: I'm interested in renting a two-bedroom apartment''}. This category is further divided into two categories \emph{word matching} and \emph{paraphrasing} in previous work~\cite{chen2016thorough,trischler2017newsqa}.

\item \textbf{Reasoning} Questions that cannot be answered by the surface meaning of a single sentence belong to this category. We further define four subcategories as follows. 
	\begin{itemize}[leftmargin=*]
    \item \textbf{Summary} Answering this kind of questions requires the whole picture of a dialogue, such as the topic of a dialogue and the relation between speakers (\eg, D$2$-Q$3$ in Table~\ref{tab:sample}). Under this category, questions such as \emph{``What are the two speakers talking about?''} and \emph{``What are the speakers probably doing?"} are frequently asked.

    \item \textbf{Logic} We require logical reasoning to answer questions in this category. We usually need to identify logically implied relations among multiple sentences in a dialogue. To reduce the ambiguity during the annotation, we regard a question that can only be solved by considering the content from multiple sentences and does not belong to the \emph{summary} subcategory that involves all the sentences in a dialogue as a \emph{logic} question. Following this definition, both D$2$-Q$1$ and D$2$-Q$2$ in Table \ref{tab:sample} belong to this category. %

    \item \textbf{Arithmetic} Inferring the answer requires arithmetic knowledge (\eg, D$2$-Q$1$ in Table \ref{tab:sample} requires $25-1+2=26$).
    
    \item \textbf{Commonsense} To answer questions under this subcategory, besides the textual information in the dialogue, we also require additional commonsense knowledge that cannot be obtained from the dialogue. For instance, all questions in Table~\ref{tab:sample} fall under this category. D$2$-Q$1$ and D$2$-Q$2$ in Table~\ref{tab:sample} belong to both \emph{logic} and \emph{commonsense} since they require multiple sentences as well as commonsense knowledge for question answering. There exist multiple types of commonsense knowledge in DREAM such as the well-known properties of a highly-recognizable entity (\eg, D$2$-Q$1$ in Table \ref{tab:sample}), the prominent relationship between two speakers (\eg, D$2$-Q$3$ in Table \ref{tab:sample}), the knowledge of or shared by a particular culture (\eg, when a speaker says \emph{``Cola? I think it tastes like medicine.''}, she/he probably means \emph{``I don't like cola.''}), and the cause-effect relation between events (\eg, D$1$-Q$1$ in Table~\ref{tab:sample1}). We refer readers to~\newcite{lobue2011types} for detailed definitions. 
    \end{itemize}
\end{itemize}

Table~\ref{tab:data:category} shows the question type distribution labeled by two human annotators on $25\%$ questions randomly sampled from the development and test sets. Besides the previously defined question categories, we also report the percentage of questions that require reasoning over multiple sentences (\ie, \emph{summary} or \emph{logic} questions) and the percentage of questions that require the surface-level understanding or commonsense/math knowledge based on the content of a single sentence. As a question can belong to multiple reasoning subcategories, the summation of the percentage of reasoning subcategories is not equal to the percentage of reasoning. The Cohen's kappa coefficient is $0.67$ on the development set and $0.68$ on the test set.

\begin{table}[htbp!]
\footnotesize
\centering
\begin{tabular}{lccc}
\toprule
\bf{Question Type}  & \bf Dev & \bf Test  &  \bf{Dev + Test} \\
\midrule
Matching & 13.0 & 10.3 & 11.7 \\
Reasoning & 87.0 & 89.7 & 88.3 \\
$~~~~~~~~$Summary & 8.4 & 15.9 & 12.1 \\
$~~~~~~~~$Logic & 74.5 & 70.4 & 72.5 \\
$~~~~~~~~$Arithmetic & 5.1 & 3.6 & 4.4 \\
$~~~~~~~~$Commonsense & 31.5 & 35.9 & 33.7 \\
\midrule
Single sentence & 17.1 & 13.7 & 15.4 \\
Multiple sentences & 82.9 &  86.3  &  84.6 \\
\bottomrule
\end{tabular}
\caption{Distribution of question types (\%).}
\label{tab:data:category}
\end{table}

Dialogues in DREAM are generally clean and mostly error-free since they are carefully designed by educational experts. However, it is not guaranteed that each dialogue is written or proofread by a native speaker. Besides, dialogues tend to be more proper and less informal for exam purposes. To have a rough estimation of the quality of dialogues in DREAM and the differences between these dialogues and more casual ones in movies or TV shows, we run a proofreading tool -- Grammarly\footnote{https://app.grammarly.com.} -- on all the dialogues from the annotated $25\%$ instances of the development set and the same size ($20.7$k tokens) of dialogues from \emph{Friends}, a famous American TV show whose transcripts are commonly used for dialogue understanding~\cite{chen2016character,ma2018challenging}. As shown in Table~\ref{tab:data:quality}, there exist fewer spelling mistakes and the overall score is slightly higher than that of the dialogues in \emph{Friends}. Based on the evaluated instances, articles and verb forms are the two most frequent grammar error categories ($10$ and $8$, respectively, out of $23$) in DREAM. Besides, the language tends to be less precise in DREAM, indicated by the number of vocabulary suggestions. For example, experts tend to use expressions such as \emph{``really hot''}, \emph{``really beautiful''}, \emph{``very bad''}, and \emph{``very important''} instead of more appropriate yet more advanced adjectives that might hinder reading comprehension of learners with smaller vocabularies. According to the explanations provided by the tool, the readability scores for both datasets fall into the same category \emph{``Your text is very simple and easy to read, likely to be understood by an average 5th-grader (age 10)''}.

\begin{table}[]
\centering
\footnotesize
\begin{tabular}{lcc}
\toprule
\bf Metric                                  & \bf DREAM            & \bf Friends  \\
\midrule
  \# of spelling errors                     &    11                &  146        \\
  \# of grammar errors                      &    23                &  16        \\
  \# of conciseness suggestions             &    6                 &  2           \\
  \# of vocabulary suggestions              &    18                &  3           \\
\midrule
  General Performance                       &    98.0                & 95.0      \\
  Readability Score                         &    93.7                &  95.3      \\
\bottomrule
\end{tabular}
\caption{Comparison of the quality of dialogues from DREAM and Friends (a TV show).} 
\label{tab:data:quality}
\end{table}

\section{Approaches}
\label{sec:methods}

We formally introduce the dialogue-based reading comprehension task and notations in Section~\ref{sec:formulation}. To investigate the effects of different kinds of general world knowledge and dialogue structure, we incorporate them into rule-based approaches (Section~\ref{sec:unsupervised}) as well as non-neural (Section~\ref{sec:gbdt}) and neural (Section~\ref{sec:transformer}) machine learning approaches. We describe in detail preprocessing and training in Section~\ref{subsec:prepAndHyper}.

\subsection{Problem Formulation and Notations}
\label{sec:formulation}

We start with a formal definition of the dialogue-based multiple-choice reading comprehension task. An $n$-turn dialogue $D$ is defined as $D=\{s_1 \colon t_1,s_2 \colon t_2,\ldots,s_n \colon t_n$\}, where $s_i$ represents the speaker ID (\eg, \emph{``M''} and \emph{``W''}), and $t_i$ represents the text of the $i^{th}$ turn. Let $Q$ denote the text of question, and $O_{1..3}$ denote the text of three answer options. The task is to choose the correct one from answer options $O_{1..3}$ associated with question $Q$ given dialogue $D$. In this paper, we regard this task as a three-class classification problem, each class corresponding to an answer option.

For convenience, we define the following notations, which will be referred in the rest of this paper. Let $D^s$ denote the turns spoken by speaker $s$ in $D$. Formally, $D^s=\{s_{i_1}\colon t_{i_1},s_{i_2} \colon t_{i_2},\ldots,s_{i_m} \colon t_{i_m}\}$ where $\{i_1,i_2,\ldots,i_m\} = \{i \, | \, s_i=s\}$ and $i_1<i_2<\ldots<i_m$. In particular, $s=*$ denotes all the speakers. $W^{D^s}$ and $W^{O_i}$ denote the ordered set of the running words (excluding punctuation marks) in $D^s$ and $O_i$ respectively. Questions designed for dialogue-based reading comprehension often focus on a particular speaker. If there is exactly one speaker mentioned in a question, we use $s_Q$ to denote this target speaker. Otherwise, $s_Q=*$. For example, given the dialogue in Table~\ref{tab:sample}, $s_Q=$\emph{``M''} for Question $1$ and $2$, and $s_Q=*$ for Question $3$. %

\subsection{Rule-Based Approaches}
\label{sec:unsupervised}

We first attempt to incorporate dialogue structure information into \textit{sliding window} (SW), a rule-based approach developed by~\newcite{richardson2013mctest}. This approach matches a bag of words constructed from a question $Q$ and one of its answer option $O_i$ with a given document, and calculates the TF-IDF style matching score for each answer option. 

Let $\hat{D}^s$, $\hat{Q}$, and $\hat{O}_{i}$ be the unordered set of distinct words (excluding punctuation marks) in $D^s$, $Q$, and $O_i$ respectively. Instead of only regarding dialogue $D$ as a non-conversational text snippet, we also pay special attention to the context that is relevant to the target speaker mentioned in the question. Therefore, given a target speaker $s_Q$, we propose to compute a \textit{speaker-focused} sliding window score for each answer option $O_i$, by matching a bag of words constructed from $Q$ and $O_i$ with $D^{s_Q}$ (\ie, turns spoken by $s_Q$). Given speaker $s$, we formally define the sliding window score $sw$ of $O_i$ as:

\begin{equation}
\small
sw_i^s=\max_{j}\sum_{k=1\dots|T_i|}
    \begin{cases}
      \text{ic}^{s}(W_{j+k}^{D^s}) & \text{if}\ W_{j+k}^{D^s}\in T_i \\
      \hfil 0 & \text{otherwise}
    \end{cases}
\label{eq:sw}
\end{equation}%
where $\text{ic}^s(w)=\log\left(1+\frac{1}{\sum_i \mathds{1}(W_{i}^{D^s}=w)}\right)$, $T_i = \hat{O}_i \cup \hat{Q}$, and $W_i^{D^s}$ denotes the $i$-th word in $W^{D^s}$. Based on the above definitions, we can regard $sw_i^{*}$ as the general score defined in the original sliding window approach, and $sw_i^{s_Q}$ represents the speaker-focused sliding window score considering the target speaker $s_Q$. %

Since sliding window score ignores long-range dependencies,~\newcite{richardson2013mctest} introduce a distance-based variation (DSW), in which a word-distance based score is subtracted from the sliding window score to arrive at the final score. Similarly, we calculate the speaker-focused distance-based score given a ($Q$, $O_i$) pair and $s_Q$, by counting the distance between the occurrence of a word in $Q$ and a word in $O_i$ in $D^{s_Q}$. More formally, given speaker $s$ and a set of stop words\footnote{We use the list of stop words from NLTK~\cite{bird2004nltk}.} $U$, the distance-based score $d$ of $O_i$ is defined as

\begin{equation}
\small
d_i^s =
    \begin{cases}
      \hfil 1 & \text{if}\ |I_Q^s|=0\ \text{or}\ |I_{O_i}^s|=0 \\
      \frac{\delta_{i}^s}{|W^{D^s}|-1} & \text{otherwise}
    \end{cases}
\label{eq:d1}
\end{equation} where $I_Q^s = (\hat{Q}\cap \hat{D}^s)-U$, $I_{O_i}^s = (\hat{O}_i\cap \hat{D}^s)-\hat{Q}-U$, and $\delta_{i}^s$ is the minimum number of words between an occurrence of a question word and an answer option word in $W^{D^s}$, plus one. The formal definition of $\delta_{i}^s$ is as follows.

\begin{equation}
\small
\delta_{i}^s = \min\limits_{W_j^{D^s}\in I_Q^s, W_k^{D^s}\in I_{O_i}^s} |j-k| + 1
\label{eq:delta}
\end{equation}

Based on the above definitions, we can regard $d_i^{*}$ as the distance-based score defined in the original sliding window approach, and $d_i^{s_Q}$ represents the speaker-focused distance-based score considering speaker $s_Q$. In addition, the final distance-based sliding window score of $O_i$~\cite{richardson2013mctest} can be formulated as
\begin{equation}
\small
sw_i^{*}-d_i^{*}
\label{eq:odsw}
\end{equation}

Compared to (\ref{eq:odsw}) that only focuses on the general (or speaker-independent) information (\ie, $sw_i^*$ and $d_i^{*}$), we can capture general and speaker-focused information (i.e. $sw_i^{s_Q}$ and $d_i^{s_Q}$) simultaneously by averaging them:
\begin{equation}
\small
\frac{sw_i^{s_Q}+sw_i^{*}}{2} - \frac{d_i^{s_Q}+d_i^{*}}{2}
\label{eq:sw+d}
\end{equation}

Since a large percentage of questions cannot be solved by word-level matching, we also attempt to incorporate general world knowledge into our rule-based method. We calculate $cs_{i}^{s}$, the maximum cosine similarity between $O_i$ and consecutive words of the same length in $W^{D^s}$, as:

\begin{equation}
\small
cs_{i}^{s} = \max_{j}\cos\left(\overline{W^{O_i}}, \overline{W^{D^s}_{j\ldots j+|W^{O_i}|-1}}\right)
\label{eq:cs}
\end{equation} where $\overline{x}$ is obtained by averaging the embeddings of the constituent words in $x$. Here we use ConceptNet embeddings~\cite{speer2017conceptnet} since they leverage the knowledge graph that focuses on general world knowledge. Following (\ref{eq:sw+d}), we capture both general and speaker-focused semantic information within a dialogue as follows.
\begin{equation}
\small
\frac{cs_{i}^{s_Q} + cs_{i}^{*}}{2}
\label{eq:cnv}
\end{equation}

To make the final answer option selection, our rule-based method combines (\ref{eq:sw+d}) and (\ref{eq:cnv}):
\begin{equation}
\small
\arg \max_i \frac{sw_i^{s_Q}+sw_i^{*}}{2} - \frac{d_i^{s_Q}+d_i^{*}}{2} + \frac{cs_{i}^{s_Q} + cs_{i}^{*}}{2}
\label{eq:dswpp}
\end{equation}

\subsection{Feature-Based Classifier}
\label{sec:gbdt}

To explore what features are effective for dialogue understanding, we first consider a gradient boosting decision tree (GBDT) classifier. Besides the conventional bag-of-words based features, we primarily focus on features related to general world knowledge and dialogue structure.

\begin{itemize}
\item \textbf{Bag of words of each answer option}.
\item \textbf{Features inspired by rule-based approaches:} we adopt the features introduced in Section~\ref{sec:unsupervised}, including speaker-independent scores (\ie, $sw_{i}^{*}$ and $d_{i}^{*}$) and speaker-focused scores (\ie, $sw_{i}^{s_Q}$ and $d_{i}^{s_Q}$).
\item \textbf{Matching position:} $p_{1..3}^{s_Q}$ and $p_{1..3}^*$, where $p_i^s$ is the last position (in percentage) of a word in $D^s$ that is also mentioned in $O_i$; $0$ if none of the words in $D^s$ is mentioned in $O_i$. We consider matching position due to our observation of the existence of concessions and negotiations in dialogues~\cite{Amgoud2007}. We assume the facts or opinions expressed near the end of a dialogue tend to be more critical for us to answer a question.

\item \textbf{Pointwise mutual information (PMI):}  $pmi_{\max,1..3}^{s_Q}$, $pmi_{\max,1..3}^*$, $pmi_{\min,1..3}^{s_Q}$, $pmi_{\min,1..3}^*$, $pmi_{\text{avg},1..3}^{s_Q}$, and $pmi_{\text{avg},1..3}^*$,
where $pmi_{f,i}^{s}$ is defined as
\begin{equation}
\small
pmi_{f,i}^{s} = \frac{\sum_{j}\log f_k \frac{C_2(W_{j}^{O_i}, W_{k}^{D^s})}{C_1(W_{j}^{O_i})C_1(W_{k}^{D^s})}}{|W^{O_i}|}
\end{equation} 

$C_1(w)$ denotes the word frequency of $w$ in external copora (we use Reddit posts~\cite{tan+lee:15}), and $C_2(w_1, w_2)$ represents the co-occurrence frequency of word $w_1$ and $w_2$ within a distance $<K$ in external copora. We use PMI to evaluate the relatedness between the content of an answer option and the target-speaker-focused context based on co-occurrences of words in external corpora, inspired by previous studies on narrative event chains~\cite{chambers2008unsupervised}.

\item \textbf{ConceptNet relations (CR):} $cr_{1..3,1..|R|}$. $R=\{r_1,r_2,\ldots\}$ is the set of ConceptNet relation types (\eg, \emph{``CapableOf''} and \emph{``PartOf''}). $cr_{i,j}$ is the number of relation triples ($w_1$, $r_j$, $w_2$) that appear in the ConceptNet~\cite{speer2017conceptnet}, where $w_1$ represents a word in answer option $O_i$, $w_2$ represents a word in $D$, and the relation type $r_j \in R$. Similar to the motivation of using PMI, we use CR to capture the association between an answer option and the source dialogue based on raw co-occurrence counts in the commonsense knowledge base.

\item \textbf{ConceptNet embeddings (CE):} besides the lexical similarity based on string matching, we also calculate $cs_{1..3}^{*}$ and $cs_{1..3}^{s_Q}$, where $cs_{i}^{*}$ and $cs_{i}^{s_Q}$ represent the maximum cosine similarity between $O_i$ and consecutive words of the same length in $D$ and $D^{s_Q}$ respectively (Expression~\ref{eq:cs} in Section~\ref{sec:unsupervised}). We use ConceptNet embeddings~\cite{speer2017conceptnet} since they leverage the general world knowledge graph. 

\end{itemize}

\subsection{End-To-End Neural Network}
\label{sec:transformer}

\begin{figure*}[!ht]
   \begin{center}
   \includegraphics[width=0.95\textwidth,height=4cm]{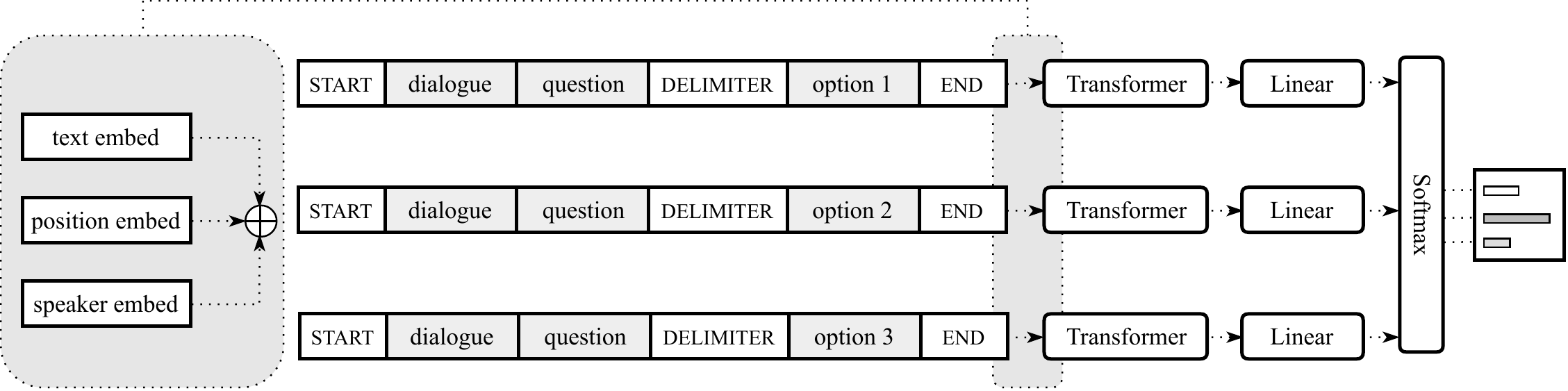}
   \end{center}
 \caption{Overall neural network framework.}
 \label{fig:method:nnStruct}
\end{figure*}

Our end-to-end neural model is based on a generative pre-trained language model (LM). We follow the framework of finetuned transformer LM (FTLM)~\cite{radfordimproving} and make modifications for dialogue-based reading comprehension.

The training procedure of FTLM consists of two stages. The first stage is to learn a high-capacity language model on a large-scale unsupervised corpus of tokens $\mathcal{U}=\{u_1,\ldots,u_n\}$ by maximizing the following likelihood:
\begin{equation}
\small
L_{LM}(\mathcal{U})=\sum_i \log P(u_i\, | \,u_{i-k},\ldots,u_{i-1};\Theta)
\end{equation}
where $k$ is the context window size, and the conditional probability $P$ is modeled by a multi-layer transformer decoder~\cite{liu2018generating} with parameters $\Theta$. In the second stage, the model is adapted to a labeled dataset $\mathcal{C}$, where each instance consists of a sequence of input tokens $x^1,\ldots,x^m$ with a label $y$, by maximizing:
\begin{equation}
\small
L(\mathcal{C})=\sum_{x,y} \log P (y\, | \,x^1,\ldots,x^m) + \lambda L_{LM}(\mathcal{C})
\end{equation}
where $P (y \, | \,  x^1,\ldots,x^m)$ is obtained by a linear $+$ softmax layer over the final transformer block's activation, and $\lambda$ is the weight for language model. For multiple-choice reading comprehension, the input tokens $x^1,\ldots,x^m$ come from the concatenation of a start token, dialogue, question, a delimiter token, answer option, and an end token; $y$ indicates if the answer option is correct. We refer readers to \newcite{radfordimproving} for more details.

Since the original FTLM framework already leverages rich linguistic information from a large unlabeled corpus, which can be regarded as a type of tacit general world knowledge, we investigate whether additional dialogue structure can further improve this strong baseline. We propose \textit{speaker embedding} to better capture dialogue structure. Specifically, in the original framework, given an input context $(u_{-k},\ldots,u_{-1})$ of the transformer, the encoding of $u_{-i}$ is $\bm{we}(u_{-i}) + \bm{pe}(i)$, where $\bm{we}(\cdot)$ is the word embedding, and $\bm{pe}(\cdot)$ is the position embedding. When adapting $\Theta$ to DREAM, we change the encoding to $\bm{we}(u_{-i}) + \bm{pe}(i) + \bm{se}(u_{-i},s_Q)$ where the speaker embedding
$\bm{se}(u_{-i},s_Q)$ is (a) $\bm 0$ if the token $u_{-i}$ is not in the dialogue (i.e. it is either a start/end/delimiter token or a token in the question/option); (b) $\bm e_{target}$ if the token is spoken by $s_Q$; (c) $\bm e_{rest}$ if the token is in the dialogue but not spoken by $s_Q$. $\bm e_{target}$ and $\bm e_{rest}$ are trainable and initialized randomly. We show the overall framework in Figure~\ref{fig:method:nnStruct}.

\subsection{Preprocessing and Training Details}
\label{subsec:prepAndHyper}

For all the models, we conduct coreference resolution to determine speaker mentions of $s_Q$ based on simple heuristics. Particularly, we map three most common speaker abbreviations (\ie, \emph{``M''}; \emph{``W''} and \emph{``F''}) that appear in dialogues to their eight most common corresponding mentions (\ie, \emph{``man''}, \emph{``boy''}, \emph{``he''}, and \emph{``his''}; \emph{``woman''}, \emph{``girl''}, \emph{``she''}, and \emph{``her''}) in questions. We keep speaker abbreviations unchanged, since neither replacing them with their corresponding full forms nor removing them contributes to the performance based on our experiments.

For the neural model mentioned in Section~\ref{sec:transformer}, most of our parameter settings follow~\newcite{radfordimproving}. We adopt the same preprocessing procedure and use their publicly released pre-trained language model, which is trained on the BooksCorpus dataset~\cite{zhu2015aligning}. We set the batch size to $8$, language model weight $\lambda$ to $2$, and maximum epochs of training to $10$. 

For other models, we use the following preprocessing steps. We tokenize and lowercase the corpus, convert number words to numeric digits, normalize time expressions to $24$-hour numeric form, and deal with negation by removing interrogative sentences that receive \emph{``no''} as the reply. We use the gradient boosting classifier implemented in the scikit-learn toolkit~\cite{pedregosa2011scikit}. We set the number of boosting iterations to $600$ and keep the rest of hyperparameters unchanged. The distance upper bound $K$ for PMI is set to $10$.

We perform several runs of machine learning models (Section~\ref{sec:gbdt} and Section~\ref{sec:transformer}) with randomness introduced by different random seeds and/or GPU nondeterminism and select the model or models (for ensemble) that perform best on the development set.

%
\section{Experiment}
\label{sec:experiment}

\subsection{Baselines}
\label{sec:baselines}

We implement several baselines, including rule-based methods and state-of-the-art neural models. 

\begin{table*}[ht!]
\footnotesize
\centering
\begin{tabular}{lcc}
\toprule
\textbf{Method} & \textbf{Dev} & \textbf{Test} \\
\midrule
Random & 32.8 & 33.4 \\
Word Matching (WM)~\cite{yih2013question} & 41.7  & 42.0   \\ 
Sliding Window (SW)~\cite{richardson2013mctest} & 42.6 &  42.5  \\
Distance-Based Sliding Window (DSW)~\cite{richardson2013mctest} & 44.4 & 44.6   \\
\midrule
Stanford Attentive Reader (SAR)~\cite{chen2016thorough} & 40.2 & 39.8  \\
Gated-Attention Reader (GAR)~\cite{dhingra2017gated}  & 40.5 & 41.3 \\
Co-Matching (CO)~\cite{wang2018co}   & 45.6 & 45.5\\
Finetuned Transformer LM (FTLM)~\cite{radfordimproving} & 55.9 & 55.5  \\
\midrule
\textsl{Our Approaches:} & & \\
DSW$++$ (DSW w/ Dialogue Structure and ConceptNet Embedding)  & 51.4  & 50.1  \\
GBDT$++$ (GBDT w/ Features of Dialogue Structure and General World Knowledge)  & 53.3 & 52.8 \\
FTLM$++$ (FTLM w/ Speaker Embedding)   & \textbf{57.6} & \textbf{57.4} \\
\midrule
Ensemble of 3 FTLM$++$  & 58.1  & 58.2 \\
Ensemble of 1 GBDT$++$ and 3 FTLM$++$  & \textbf{59.6}  & \textbf{59.5} \\ %
\midrule
\midrule
Human Performance & 93.9$^\star$ & 95.5$^\star$ \\
Ceiling Performance & 98.7$^\star$  & 98.6$^\star$  \\
\bottomrule
\end{tabular}
\caption{Performance in accuracy ($\%$) on the DREAM dataset. Performance marked by $\star$ is reported based on 25\% annotated questions from the development and test sets.} %
\label{tab:eval:performance}
\end{table*}

\begin{itemize}

\item \textbf{Word Matching} This strong baseline~\cite{yih2013question} selects the answer option that has the highest count of overlapping words with the given dialogue.

\item \textbf{Sliding Window} We implement the sliding window approach (\ie, $\arg \max_i sw_i^*$) and its distance-based variation (\ie, $\arg \max_i sw_i^*-d_i^*$)~\cite{richardson2013mctest} introduced in Section~\ref{sec:unsupervised}. 

\item \textbf{Enhanced Distance-Based Sliding Window (DSW$++$)} We also use general world knowledge and speaker-focused information to improve the original sliding window baseline, formulated in Expression~\ref{eq:dswpp} (Section~\ref{sec:unsupervised}). 

\item \textbf{Stanford Attentive Reader} This neural baseline compares each candidate answer (\ie, entity) representation to the question-aware document representation built with attention mechanism~\cite{hermann2015teaching,chen2016thorough}.~\newcite{lai2017race} adds a bilinear operation to compare document and answer option representations to answer multiple-choice questions. 

\item \textbf{Gated-Attention Reader} The baseline models multiplicative question-specific document representations based on a gated-attention mechanism~\cite{dhingra2017gated}, which are then compared to each answer option~\cite{lai2017race}. 

\item \textbf{Co-Matching} This state-of-the-art multiple-choice reading comprehension model explicitly treats question and answer option as two sequences and jointly matches them against a given document~\cite{wang2018co}.

\item \textbf{Finetuned Transformer LM} This is a general task-agnostic model introduced in Section~\ref{sec:transformer}, which achieves the best reported performance on several tasks requiring multi-sentence reasoning~\cite{radfordimproving}. %
\end{itemize}

We do not investigate other ways of leveraging pre-trained deep models such as adding ELMo representations~\cite{peters2018deep} as additional features to a neural model since recent studies show that directly fine-tuning a pre-trained language model such as FTLM is significantly superior on multiple-choice reading comprehension tasks~\cite{radfordimproving,chen2018convolutional}. We do not apply more recent extractive models such as AOA~\cite{cui2017attention} and QANet~\cite{yu2018qanet} since they aim at precisely locating a span in a document. When adapted to solve questions with abstractive answer options, extractive models generally tend to perform less well~\cite{chen2016thorough,dhingra2017gated,lai2017race}.

\subsection{Results and Analysis}

We report the performance of the baselines introduced in Section~\ref{sec:baselines} and our proposed approaches in Table~\ref{tab:eval:performance}. We report the averaged accuracy of two annotators as the human performance. The proportion of valid questions (\ie, an unambiguous question with a unique correct answer option provided) that are manually checked by annotators on the annotated test and development sets is regarded as the human ceiling performance. 

\textbf{Surface matching is insufficient.} Experimental results show that neural models that primarily exploit surface-level information (\ie, SAR, GAR, and CO) attain a performance level close to that of simple rule-based approaches (\ie, WM, SW, and DSW). The highest accuracy achieved by CO is $45.5\%$, a similar level of performance to the rule-based method DSW ($44.6\%$).  

\textbf{It is helpful to incorporate general world knowledge and dialogue structure.}
We see a significant gain $5.5\%$ in accuracy when enhancing DSW using general world knowledge from ConceptNet embeddings and considering speaker-focused information (Section~\ref{sec:unsupervised}). FTLM, which leverages rich external linguistic knowledge from thousands of books, already achieves a much higher accuracy $55.5\%$ compared to previous state-of-the-art machine comprehension models, indicating the effectiveness of general world knowledge. Experimental results show that our best single model FTLM$++$ significantly outperforms FTLM (p-value $=0.03$), illustrating the usefulness of additional dialogue structure. Compared to the state-of-the-art neural reader Co-Matching that primarily explore surface-level information ($45.5\%$), the tacit general world knowledge (in the pre-trained language model) and dialogues structure in FTLM$++$ lead to an absolute gain of $11.9\%$ in accuracy.

\textbf{Ensembling different types of methods can bring further improvements.} We employ the majority vote strategy to obtain the ensemble model performance. While GBDT$++$ ($52.8\%$) itself does not outperform FTLM$++$, GBDT$++$ can serve as a supplement to FTLM$++$ as they have diverse types of general world knowledge and model architectures. We achieve the highest accuracy $59.5\%$ by ensembling one GBDT$++$ and three FTLM$++$.

\subsection{Ablation Tests}
We conduct ablation tests to evaluate the individual components of our proposed approaches (Table~\ref{tab:eval:ablation}). In Table~\ref{tab:eval:feature_types}, we summarize the involved types of dialogue structure and general world knowledge in our approaches. 

\noindent\textbf{Dialogue Structure} Specifically, we observe $1.4\%$ drop in accuracy if we set the target speaker $s_Q$ to $*$ for all questions when we apply DSW$++$. We observe a similar performance drop when we remove speaker-focused features from GBDT$++$. In addition, removing speaker embeddings from FTLM$++$ leads to $1.7\%$ drop in accuracy (in this case, the model becomes the original FTLM). These results consistently indicate the usefulness of dialogue structure for dialogue understanding. 

\noindent\textbf{General World Knowledge} We also investigate the effects of general world knowledge. The accuracy of DSW$++$ drops by $4.7\%$ if we remove ConceptNet embeddings (CE) by deleting the last term of Expression~\ref{eq:dswpp} in Section~\ref{sec:unsupervised}. Additionally, the accuracy of GBDT$++$ drops by $6.2\%$ if we remove all the general world knowledge features (\ie, ConceptNet embeddings/relations and PMI), leading to prediction failures on questions such as \emph{``What do we learn about the man?''} whose correct answer option \emph{``He is health-conscious.''} is not explicitly mentioned in the source dialogue \emph{``\textbf{M}: We had better start to eat onions frequently, Linda. \textbf{W}: But you hate onions, don't you? \textbf{M}: Until I learned from a report from today's paper that they protect people from flu and colds. After all, compared with health, taste is not so important.''}. Moreover, if we train FTLM$++$ with randomly initialized transformer weights instead of weights pre-trained on the external corpus, the accuracy drops dramatically to $36.2\%$, which is only slightly better than a random baseline.

\begin{table}[h!t]
\footnotesize
\begin{tabular}{p{1.3cm}p{2.3cm}p{2.8cm}}
\toprule
       & General World $~$Knowledge                        & Dialogue Structure\\
\midrule
DSW$++$  & CE                                             & speaker-focused $~$scores \\
\midrule
GBDT$++$ & CE, CR, and PMI                                    & speaker-focused $~$features \\
\midrule
FTLM$++$ & pre-trained LM                                  & speaker embedding     \\
\bottomrule
\end{tabular}
\caption{Types of general world knowledge and dialogue structure investigated in our approaches.} 
\label{tab:eval:feature_types}
\end{table}

\begin{table}[h!t]
\centering
\footnotesize
\begin{tabular}{lcc}
\toprule
\textbf{Method} & \textbf{Accuracy}   &  $\bf \Delta$ \\
\midrule
DSW$++$ & 51.4  & $-$ \\
\ $-$ dialogue structure & 50.0 & -1.4\\
\ $-$ CE & 46.7 & -4.7 \\
\midrule
GBDT$++$ & 53.3  & $-$ \\
\ $-$ bag of words & 51.6 & -1.7       \\
\ $-$ rule-based features & 51.2  & -2.1  \\
\ $-$ matching position & 53.0 & -0.3  \\
\ $-$ dialogue structure & 51.9 & -1.4 \\
\ $-$ PMI & 51.4 & -1.9 \\
\ $-$ CR & 52.7 & -0.6\\
\ $-$ CE & 52.7 & -0.6\\
\ $-$ PMI, CR, CE & 47.1 & -6.2\\
\midrule
FTLM$++$ & 57.6 & $-$ \\
\ $-$ speaker embedding & 55.9 & -1.7\\
\ $-$ LM pre-training & 36.2 & -21.4\\
\bottomrule
\end{tabular}
\caption{Ablation tests on the development set (\%). Minus ($-$) indicates percentage decrease.} 
\label{tab:eval:ablation}
\end{table}

\subsection{Error Analysis}

\begin{figure}[!h]
   \begin{center}
   \includegraphics[width=0.4\textwidth,height=5cm]{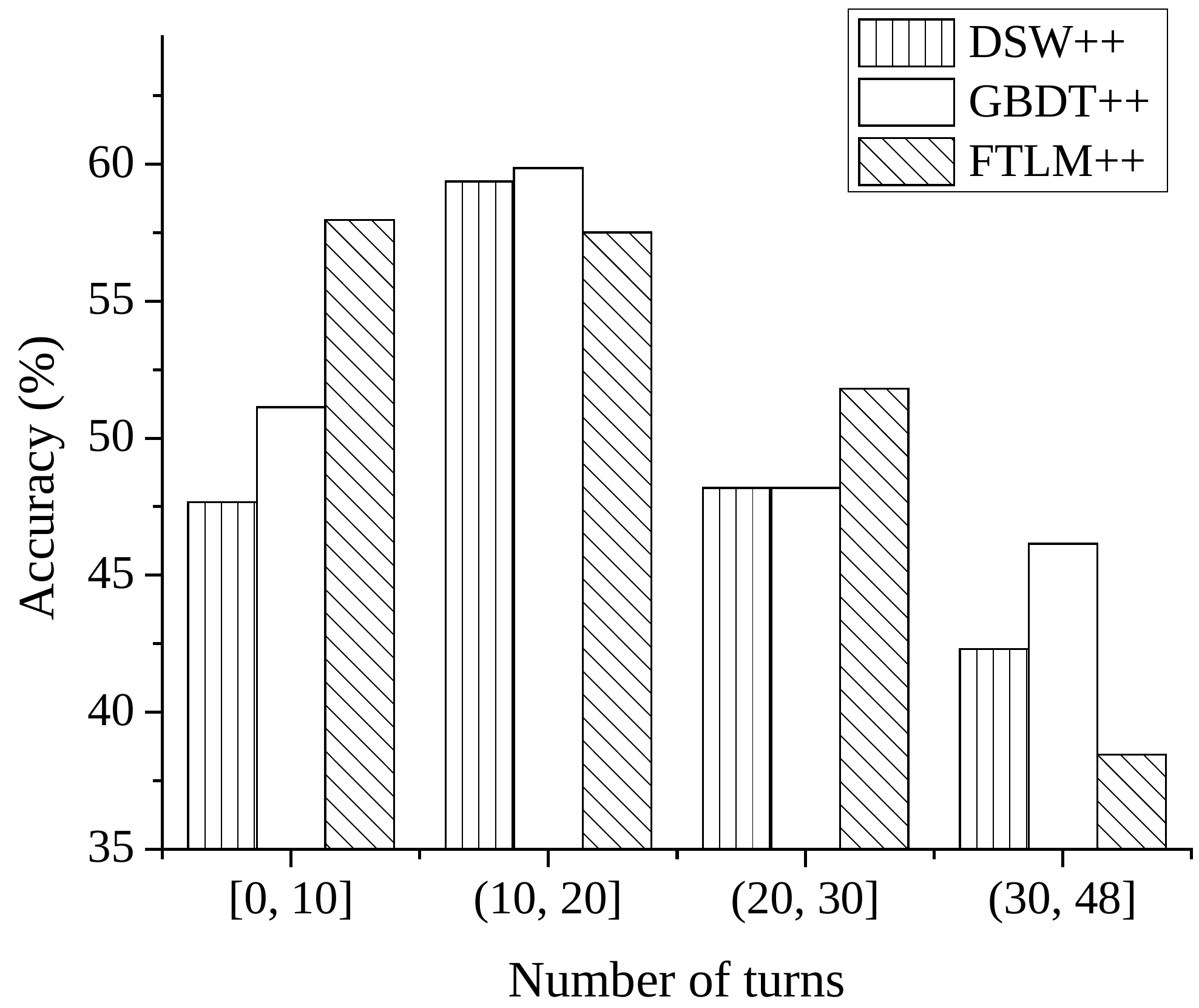}
   \end{center}
 \caption{Performance comparison of different number of turns on the test set.}
 \label{fig:eval:turn}
\end{figure}

\textbf{Impact of Longer Turns}
The number of dialogue turns has a significant impact on the performance of FTLM$++$. As shown in Figure~\ref{fig:eval:turn}, its performance reaches the peak while the number of turns ranges from $0$ to $10$ while it suffers severe performance drops when the given dialogue contains more turns. Both DSW$++$ ($56.8\%$) and GBDT$++$ ($57.4\%$) outperform FTLM$++$ ($55.7\%$) when the number of turns ranges from $10$ to $48$. To deal with lengthy context, it may be helpful to first identify relevant sentences based on a question and its associated answer options rather than using the entire dialogue context as input.

\noindent \textbf{Impact of Confusing Distractors}
For $54.5\%$ of questions on the development set, the fuzzy match score~\cite{sikes2007fuzzy} of at least one distractor answer option against the dialogue is higher than the score of the correct answer option. For questions that all models (\ie, DSW$++$, GBDT$++$, and FTLM$++$) fail to answer correctly, $73.0\%$ of them contain at least one such confusing distractor answer option. The causes of this kind of errors can be roughly divided into two categories. First, the distractor is wrongly associated with the target speaker/s mentioned in the question (\eg, answer option A and C in D$2$-Q$3$ in Table \ref{tab:sample}). Second, although the claim in the distractor is supported by the dialogue, it is irrelevant to the question (\eg, D$1$-Q$1$-B in Table \ref{tab:sample1}). A promising direction to solve this problem could be the construction of speaker-focused event chains~\cite{chambers2008unsupervised} and dialogue-specific coreference resolution systems for more reliable evidence collection in a dialogue. 

\noindent \textbf{Impact of Question Types} We further report the performance of the best single model FTLM$++$ and the GBDT$++$ baseline on the categories defined in Section~\ref{analysis} (Table~\ref{tab:eval:type}). Not surprisingly, both models perform worse than random guessing on math problems. While most of the problems can be solved by one single linear equation, it is still difficult to apply recent neural math word problem solvers~\cite{huang2018neural,leiw2018} due to informal dialogue-based problem descriptions and the requirement of commonsense inference. For example, given the dialogue:\\
\emph{``\textbf{W}: The plane arrives at 10:50. It is already 10:40 now. Be quick! \textbf{M}: Relax. Your watch must be fast. There are still twenty minutes left.''}, \\we need prior knowledge to infer that the watch of the man is showing incorrect time $10$:$40$. Instead, $10$:$50$ should be used as the reference time with the time interval \emph{``twenty minutes left''} together to answer the question \emph{``What time is it now?''}.

Results show that GBDT$++$ is superior to the fine-tuned language model on the questions under the category \emph{matching} ($68.1\%$ \vs $57.0\%$) and the latter model is more capable of answering implicit questions (\eg, under the category \emph{summary}, \emph{logic}, and \emph{commonsense}) which require aggregation of information from multiple sentences, the understanding of the entire dialogue, or the utilization of world knowledge. Therefore, it might be useful to leverage the strengths of individual models to solve different types of questions.

\begin{table}[htbp!]
\footnotesize
\centering
\begin{tabular}{lccc}
\toprule
 \bf{Question Type}    &  \bf{FTLM$++$}  &  \bf{GBDT$++$}  \\
\midrule 
Matching               &   57.0     &  \bf 68.1    \\
Reasoning              &   \bf 56.8     &  49.4    \\
$~~~~~~~~$Summary      &   \bf 73.6     &  47.1    \\
$~~~~~~~~$Logic        &   \bf 55.0     &  49.7    \\
$~~~~~~~~$Arithmetic   &   \bf 30.2     &  24.5    \\
$~~~~~~~~$Commonsense  &   \bf 53.4     &  41.7     \\
\midrule
Single sentence        &   56.5         &  \bf 63.3   \\
Multiple sentences     &   \bf 56.9     &  49.5   \\
\bottomrule
\end{tabular}
\caption{Accuracy (\%) by question type on the annotated development subset.}
\label{tab:eval:type}
\end{table}

\section{Conclusion and Future Work}

We present DREAM, the first multiple-choice dialogue-based reading comprehension dataset from English language examinations. Besides the multi-turn multi-party dialogue context, $85\%$ of questions require multiple-sentence reasoning, and $34\%$ of questions also require commonsense knowledge, making this task very challenging. We apply several popular reading comprehension models and find that surface-level information is insufficient. We incorporate general world knowledge and dialogue structure into rule-based and machine learning methods and show the effectiveness of these factors, suggesting a promising direction for dialogue-based reading comprehension. For future work, we are interested in problem generation for dialogues and investigating whether it will lead to more gains to pre-train a deep language model such as FTLM over large-scale dialogues from movies and TV shows instead of the BookCorpus dataset~\cite{zhu2015aligning} used by previous work~\cite{radfordimproving}.

\section*{Acknowledgments}
We would like to thank the editors and anonymous reviewers for their helpful feedback. 
We also thank Hai Wang from Technological Institute at Chicago for useful discussions and valuable comments.

\bibliography{tacl}
\bibliographystyle{acl_natbib}

\clearpage
\appendix

\end{document}